\documentclass[11pt]{article}

\usepackage{float}
\usepackage{emnlp2023}

\usepackage{times}
\usepackage{latexsym}
\usepackage{enumitem}
\usepackage{booktabs}
\usepackage{subfig}

\usepackage[T1]{fontenc}
\usepackage[utf8]{inputenc}

\usepackage{graphicx}

\usepackage{microtype}

\usepackage{gb4e}
\noautomath

\usepackage{dblfloatfix}
\usepackage{xcolor}

\newcommand\blfootnote[1]{%
  \begingroup
  \renewcommand\thefootnote{}\footnote{#1}%
  \addtocounter{footnote}{-1}%
  \endgroup
}

\title{Intelligent Assistant Language Understanding On Device}

\author{Cecilia Aas, {\bf Hisham Abdelsalam}, {\bf Irina Belousova}, {\bf Shruti Bhargava},
{\bf Jianpeng Cheng}, \\
{\bf Robert Daland}, {\bf Joris Driesen}, {\bf Federico Flego}, {\bf Tristan Guigue},  {\bf Anders Johannsen}, \\ {\bf Partha Lal}
{\bf Jiarui Lu}, {\bf Joel Ruben Antony Moniz},  {\bf Nathan Perkins},  {\bf Dhivya Piraviperumal}, \\ 
{\bf  Stephen Pulman}, {\bf Diarmuid Ó Séaghdha},  {\bf David Q. Sun},  {\bf John Torr},  \\
{\bf Marco Del Vecchio},  {\bf Jay Wacker},  {\bf Jason D. Williams},  {\bf Hong Yu} \\
Apple \\
\vspace*{-0.05in}
\small \{caas,
hisalam,
ibelousova,
shruti\_bhargava,
jianpeng.cheng,
rdaland,
joris\_driesen,
fflego,
tguigue,
ajohannsen,\\
\vspace*{-0.05in}
\small
partha\_lal,
jiarui\_lu,
joelrubenantony\_moniz,
nathan\_perkins,
dhivyaprp,
spulman,
doseaghdha,
dqs,
jtorr,\\
\small
mdelvecchio,
jwacker,
jason\_williams4,
hong\_yu\}@apple.com
}

\begin{document}

\maketitle 

\begin{abstract}

It has recently become feasible to run personal digital assistants on phones and other personal devices. In this paper we describe a design for a natural language understanding system that runs on device. In comparison to a server-based assistant, this system is more private, more reliable, faster, more expressive, and more accurate. We describe what led to key choices about architecture and technologies. For example, some approaches in the dialog systems literature are difficult to maintain over time in a deployment setting. We hope that sharing learnings from our practical experiences may help inform future work in the research community.\blfootnote{Authors listed in alphabetical order}

\end{abstract}

\section{Introduction}

Personal digital assistants allow users to control their electronic devices through speaking. A broad perspective of the field over the past decade or so would include fundamental research on task-oriented dialog systems (e.g. \cite{Young2013POMDPBasedSS}), commercial assistants that run on phones or other devices \cite{kollar-etal-2018-alexa}, and academic efforts to explore new design ideas  \cite{almond2017}. For researchers it is a constant challenge to find the balance between the fast pace of emerging technology with its exciting prospect of delivering ever richer functionality, and the reality that turning a design into a robust functioning system  is neither quick nor easy. In this paper we describe an NLU system that is suitable for a modern assistant, beginning with some core principles that constrained our search through the design space of possible NLU systems.

\textbf{Privacy.} \textit{The assistant can execute exclusively on the user's device; it only connects to a server when necessary.} 
Common tasks, such as messaging and reminder-taking, require assistant access to private information such as user contacts. However, it is possible to avoid exposing this information in server requests by handling such personal requests on device. Minimizing the need for a network connection can also improve speed and reliability. Note that there are some requests which must still be handled through servers. Some queries require accessing databases that are too large to fit on device; examples include knowledge queries and accessing nonlocal music catalogs. Some queries require information which updates frequently or unpredictably, such as queries about the weather and sports events.

\textbf{Modularity.} \textit{The NLU system should minimize the amount of contextual information it reasons about, and be unconstrained by the behaviour of downstream components.} 
Academic research on dialog systems finds benefits in ``end-to-end'' conversation management \cite{mrksic-etal-2017-neural,ham-etal-2020-end, wen-etal-2017-network},  whereby a single statistical component has broad control over how the assistant understands context and decides what to do next. An alternative philosophy is to find the minimal amount of conversational intelligence needed to support natural multi-turn interaction. A key motivation for this alternative is data maintenance at scale. It is very difficult even for trained annotators to reliably label what an assistant should do in the presence of rich context; moreover, policy changes can force reannotation, which does not scale well. An additional motivation is that full-featured assistants support heterogeneous experiences built by different developer teams, and there may not be a one-size-fits-all approach to dialog management.

\textbf{Conversational.} \textit{Multiturn conversations should feel natural for users, and the assistant should not be limited to understanding simple requests.} 
Here is an example of a bad interaction: \begin{enumerate}[noitemsep,nolistsep]
    \item[] \textit{User}: Set a 2 pm alarm
    \item[] \textit{Assistant}: I've set your alarm
    \item[] \textit{User}: Call it ``study time''
    \item[] \textit{Assistant}: What would you like to update?
\end{enumerate} The assistant failed to understand that the antecedent of ``it'' is precisely the alarm that the assistant just mentioned. The contextual representation must be rich enough to support reference resolution in such cases.

Many assistants parse user requests into a flat ``slot-value pair'' meaning representation. This can preclude understanding complex requests such as ``send a message to all the attendees in my next meeting'' or ``turn the volume up and resume the music''. Support for compositional and compound queries is a core design principle of our meaning representation (see also \citet{kollar-etal-2018-alexa, gupta-etal-2018-semantic-parsing,meng2022lexicon}).

In the remainder of this paper, we describe some technical solutions which are consistent with the design principles just discussed, beginning with the utterance embedding.

\section{Embeddings on device}

\label{embeddings}

A common approach for semantically-rich embeddings is to combine unsupervised pretraining with supervised fine-tuning on the target task. In an on-device setting, this does not scale well as there may be multiple, diverse consumers. Instead, a generic embedding model is used to map tokens to context-aware vectors; these are then passed on to the relevant consumers. Resource consumption is minimized with a combination of hyperparameter optimization, distillation, and quantization.

\textbf{Architecture and Training}. The base embedding closely follows the BERT architecture \cite{devlin2019bert}. The training set consists of opt-in user queries, which are mainly short and single-turn utterances. Therefore, the auxiliary task of next-sentence prediction is not suitable; instead, domain prediction is used as an auxiliary task. For the training set, we downsample shorter utterances to promote more diverse training data.

\textbf{Hyperparameter optimization}. The second-to-last layer is used as the embedding vectors, as this captures transferrable semantics \cite{rogers-etal-2020-primer,liu-etal-2019-linguistic}, and reduces the on-device computation. The input tensor is padded (or for long utterances, compressed) to a fixed width, as this yields faster runtime inference.

\textbf{Distillation and Quantization}. Distillation is a model compression technique that transfers the knowledge of a large ``teacher'' network, by training a ``student'' network to mimic the teacher. Using Tiny-BERT attention-based distillation \cite{jiao-etal-2020-tinybert}, and quantizing to 16-bit floating points (from 32-bit in training), we achieved a reduction of 20x from a full-sized model, without statistically significant loss in e2e accuracy.

\section{Featurization}
\label{spanization}

While models pre-trained on large data may learn semantic distinctions that are useful for interpreting user queries, they do not capture some aspects of user meaning. These include \textbf{personal vocabulary} such as the names of contacts and user music playlists; \textbf{technical vocabulary} which may be specific to a particular device type (such as ``light'', which may be a device's camera light or a smart light bulb); and  \textbf{contextual references}.

Personalised and technical meanings are captured by ``spans'' -- a data structure which associates some metadata with a substring of the utterance. For example, the user mention ``mom'' might be encoded with the label \texttt{personRelation}). Spans may also associate the user mention to a canonical identifier for an item in the entity store, or a subtree which encodes the item's meaning. A notable challenge for span matching is that user mentions may not  match the canonical reference form. For example, in Russian the dictionary form is \textit{papa}; but when Russians ask to call papa they use the accusative \textit{pape}. As another example, if a user has programmed a voice command ``My Morning Routine'', they may request ``Morning routine'' (no ``My'') and expect the voice command to be executed. In practice, most mismatches are handled through a combination of lemmatization and fuzzy matching, with constraints to prevent overmatching on short strings.

Contextual references are complex, and therefore treated separately in the next section.

\section{Contextual references}
\label{rewriting}

Speakers naturally refer to entities and predicates from previous turns using pronouns, ellipsis, and other contextual references. \begin{enumerate}[noitemsep,nolistsep]
    \item[] \textit{User (first turn)}: How old is Barack Obama?
    \item[] \textit{User (followup)}: What about his wife?
\end{enumerate}
When users are looking at a screen, they may refer to entities in various ways, e.g. visual position on the screen (\textit{the bottom number}) or possessor (\textit{the dentist's office}). Although such cases are heterogeneous, the common point is that reference resolution requires relevant linguistic, visual, and device state context. Our design enables contextual reference resolution through two distinct mechanisms. 

Span-based reference resolution (SR) links a referring expression to an entity from the context. It is backed by an ``entity store'', which is populated through entity pullers (which extract salient linguistic, visual, and device state context) as well as vocabulary donations from device apps. Processing begins with a low-precision high-recall mention detector step, which uses heuristics and pattern matching to ground user mentions in selected use cases (e.g. \textit{second from the bottom} in a list context). A slower but more expressive mention resolver system based on \citet{MAttNet} attempts to match user mentions against screen context using the utterance embedding.

Query rewrite (QR) replaces references and elided content with the referent, such that the rewritten query can be understood without conversational context. QR is backed by an encoder-decoder architecture that takes as input the current user utterance, previous user utterances, and assistant responses; the output is a rewrite of the current utterance \cite{tseng-etal-2021-cread,rastogi-etal-2019-scaling}. In the example above, QR replaces \textit{What about his wife} with \textit{How old is Barack Obama's wife}. This rewrite can be understood by a dedicated knowledge system without multiturn context. QR is also an effective way to handle corrections like the  following: \begin{enumerate}[noitemsep,nolistsep]
    \item[] \textit{User (first turn)}: How old is Emma Watson?
    \item[] \textit{User (followup)}: I meant Emily Watson
\end{enumerate}

 
Query rewrite and span-based reference resolution are  complementary contextual reference strategies. SR associates user mentions with canonical identifiers, which allow downstream components to retrieve necessary information from the entity store. QR replaces contextual mentions with their antecedents, in effect preprocessing the utterance for downstream systems with limited contextual resolution. In the next sections we turn to how contextual information and other features are incorporated in the meaning representation.

\section{Dialog state}

One approach to the interpretation of dialog is to see each utterance turn as creating or updating a dialog state, which tracks the history of the conversation so far. To create the next state, the input is the previous state, and any new user or system utterances, as for example in  \cite{mrksic-etal-2015-multi,henderson-etal-2014-word}. To handle ambiguity and uncertainty it is possible to maintain a distribution over alternative dialog states, and choose the most likely next state \cite{young-2014-keynote}. 

Near the opposite end of the scale is handling things one turn at a time. Modeling conversation as a pair of user input and system response is often described as a ``conversational move'' or ``dialog game'' \cite{carletta-etal-1997-reliability}. However, within such a turn-based framework it can be difficult to deal with revisions or changes, or to interpret partial or context-dependent utterances (``no, swap the
departure and arrival cities''). A number of academic studies have found that the dialog state tracking approach is generally more accurate than turn based approaches in terms of measures like task completion or intent recognition, and it is probably true to say that state-tracking is currently the dominant approach to task-based dialog (for a recent survey see \citet{balaraman-etal-2021-recent}).

Nevertheless,  we opted for a position nearer the
turn-based end of the scale, for a number of reasons. First, in the context of task-based systems (as opposed to chatbots) it is not usually necessary to keep a long dialog history. Second, some of the relevant components of dialog state that would need to be tracked are those to do with actions of the system. These include  the previous system response, what the user can see on the screen, which state the system is in. They are dependent on software components rather than the state of the dialog, and change over time, which often means that dialog state training data needs to be re-built for re-training. In general such a close coupling between code carrying out system actions and the overall state model makes maintenance very challenging.

Thus we restricted the input features for NLU to include only the previous system action, and some necessary system information, while ignoring earlier dialog state, information about the type of device and so on. The contextual reference system described in \ref{rewriting} turns out to be sufficient for many use cases.

\section{Meaning representation}

In a ``flat'' meaning representation, semantic
interpretation consists of finding values for attributes (``slot-filling''). Such a representation, though easy to process accurately,  makes it clumsy or
impossible to deal with even simple types of compositionality, such as conjunction of multiple intents in an utterance.

We developed a hierarchical meaning representation, capturing domains, verbs, attributes and values within a
rooted graph grounded in a general ontology. We annotated a corpus of around 27k conversations such as:

\noindent
{\bf User:} "Hi can you book me a flight to Paris"

\hspace*{0.2in}{\bf Flight.book}(to={\bf Location}("Paris"))

\noindent
{\bf System:} "Sure, when and where will you depart?"

\hspace*{0.2in}{\bf Prompt(Flight.book}(from=?,departingAt=?)\\
\hspace*{0.6in})

\noindent
{\bf User:} "Tomorrow from London"

\hspace{0.2in}{\bf Flight.book}(from=Location("London"),\\
\hspace*{0.4in} to={\bf Location}("Paris"),\\
\hspace*{0.4in} departingAt={\bf DateTime}(date=Tomorrow)
\hspace*{0.6in}	    )

To compare against flat representations,
we implemented an encoder-decoder model which encodes the current user utterance and limited dialog history and uses them to make a
conditional prediction of the next state, represented as a tree. This work and the corpus are more fully described in \cite{cheng-etal-2020-conversational}.
We compared the result  (exact match of
tree averaged over all
turns in the test set)  with a number of
existing slot-filling implementations. For these systems we trained and tested on a version of our corpus where
hierarchical representations had been ``flattened'' into an attribute-value equivalent. We also compared our own system
trained on this flat representation to the vanilla hierarchical version. Results are in Table~\ref{fig: accuracy}. This suggests that we can use a more expressive representation with no loss, in fact a gain in accuracy.

\begin{table}[!htb]
\begin{center}

\begin{tabular}{lr}
  \toprule
  Decoder &  Accuracy \\
  \midrule
  COMER \cite{ren-etal-2019-scalable} &  50.9\% \\
  TRADE \cite{wu-etal-2019-transferable} & 51.3\% \\
  Hierarchical, flattened & 53.5\% \\
  Hierarchical, original & 62.2\%\\
  \bottomrule
\end{tabular}
\end{center}
\caption{\label{fig: accuracy} Exact match accuracy of
meaning representation}
\end{table}

\section{Parsing}

\begin{figure*}
\includegraphics[width=\textwidth]{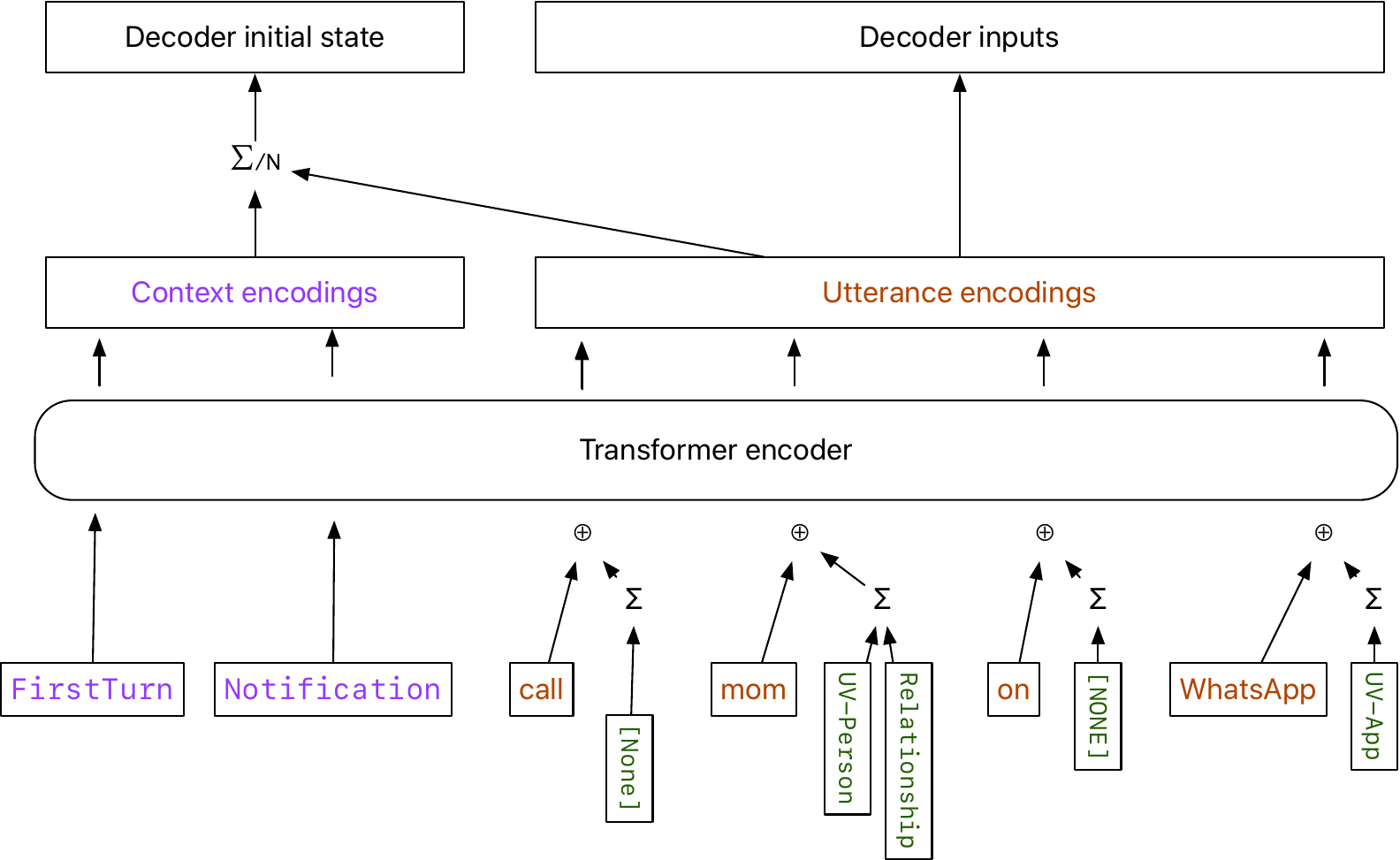}
\caption{\textbf{Encoder}. Embeddings for utterance tokens are first concatenated with span tokens along the embedding dimension, then the results are concatenated with context embeddings and fed to a transformer encoder. The average of the all transformer outputs are used as the initial hidden state of the decoder, while the utterance part of the output are sliced off and used one by one as inputs to the decoder.}
\label{fig-encoder-inputs}
\end{figure*}

The natural language understanding task for our dialogue system is best described as contextual semantic parsing \citep{Li2020ContextDS}. The parsing algorithm must balance the need for an expressive meaning representation with strict on-device memory and latency constraints. 

\textbf{Input.} The parse is conditioned on the latest user utterance and the latest system action if available, and is also sensitive to device state such as recent notifications as well as user-specific concepts (Section \ref{spanization}). Figure \ref{fig-encoder-inputs} shows how these signals are mixed and encoded. The system action is featurized using a small set of context tokens. User-specific concepts from the user utterance are embedded and concatenated to the relevant input tokens. 

\textbf{Output.} The output of the model is a sequence of tokens that can be deterministically transformed to a tree. The tree is predicted path by path, and each path is predicted roughly node by node. For the utterance ``Please create fishing trip alarm on Sundays'', the decoder predictions are shown in Table \ref{table-alarm-predictions}, while the correct tree is as follows:

{\bf Alarm.create}(\\
\hspace*{0.3in}name="Fishing trip",\\
\hspace*{0.3in}recurrence={\bf DateTime}(dayOfWeek=Sunday)\\
\hspace*{0.4in})

\begin{table}[t] \begin{tabular}{ll}
  \toprule
  Current word &  Prediction \\
  \midrule
  Please  &  \texttt{Alarm.create} \\
  {}      &  \textsc{Next} \\
  create  &  \textsc{Next} \\
  fishing &  \textsc{Copy} \\
  {} &       \textsc{Next} \\
  trip &     \texttt{Alarm.name}  \\
  {} &       \textsc{Next} \\  
  alarm &    \textsc{Next} \\
  on &       \textsc{Next}\\
  Sundays &  \texttt{Alarm.recurrence} \\
  {} &       \texttt{DateTime.dayOfWeek} \\
  {} &       \texttt{DayOfWeek.Sunday}\\
  {}      &  \textsc{End} \\  
  \bottomrule
\end{tabular}
\caption{Decoder predictions. The verb \texttt{Alarm.create} and any unaligned paths are predicted at the initial token. The \textsc{Copy} operation marks the start of the payload ''fishing trip`. It is terminated by \texttt{Alarm.name} and attached at the end of this path as a leaf node. In the training data the path \texttt{Alarm.recurrence} + \texttt{DateTime.dayOfWeek} +  \texttt{DayOfWeek.Sunday} was heuristically aligned to ``Sundays'' causing it to be output at this point. Finally decoding stops when the \textsc{End} symbol is predicted. 
}
\label{table-alarm-predictions} 
\end{table}

\textbf{Algorithm.} The model consists of an encoder-decoder pair. The encoder architecture was shown in figure \ref{fig-encoder-inputs}. Processing time is reduced by placing the majority of the parameters and computation in the encoder, keeping the recurrent decoder comparatively lightweight. Left-to-right processing and path-based tree representation yield compact output sequences. This path-based tree linearization does have some redundancy in the form of duplicated path fragments, but popular alternatives (like \citet{Choe2016ParsingAL}) have similar drawbacks. The parser avoids the use of expensive components like general purpose attention between the decoder and the encoder and a generic copy mechanism.

Assembling a tree from the predictions is done by inserting nodes incrementally, reusing nodes when they already exist in the tree. To support multiple entities of the same type for a repeatable slot (e.g. the two recipients in "Send a message to Eugene and Mark"), there is a \textsc{Flush} command which marks nodes under (any) repeatable slots as ineligible for reuse.

To produce string-valued leaf nodes in the tree we use a type of restricted copy mechanism. A \textsc{Copy} symbol marks the start of the string and subsequent \textsc{Next} predictions extend the string to the right. The copied string is terminated when a new path is predicted. The copy mechanism is constrained so that strings in the meaning representation can only be contiguous slices of the input utterance. Copying can also be performed on subtrees associated with a given substring through an entity span (see \ref{spanization}).

\begin{figure*}[t]
\centering
\includegraphics[scale=0.215]{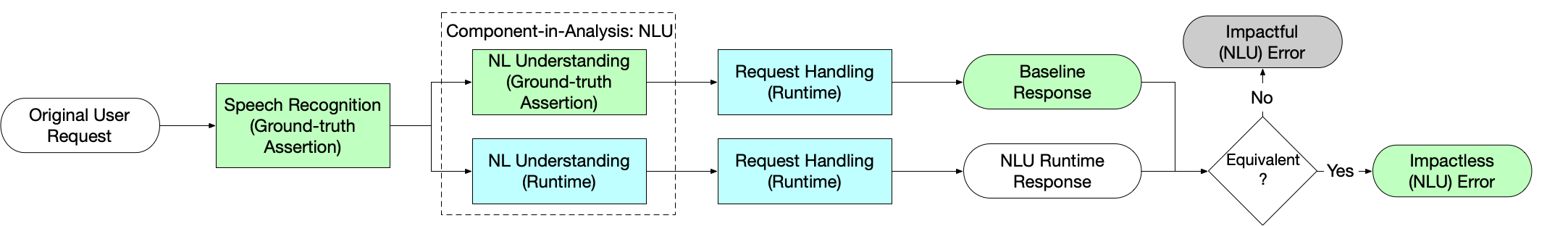} 
\caption{Holistic Evaluation Pipeline
}
\label{fig-holistic-eval}
\end{figure*}

The general parser is sufficiently expressive to cover personal requests that can be handled on device (sms, reminders, etc..). In the next section, we discuss additional parsers.

\section{Additional parsers}

Besides the general parser for personal requests, on-device understanding consists of a parser for knowledge queries (which generally require handling via a knowledge base that is too large to fit on device). The system design allows for other, specialized parsers to be added without significant refactoring. For example, a dedicated overrides parser supports parallel development of understanding and execution: overrides are defined for `hero use-cases' to support execution, while broader annotation/modeling support is brought up for the understanding system.

\section{Evaluation}

An expressive ontology introduces several challenges in evaluation. First, richer ontologies make it harder to converge on (and abide by) consistent labeling conventions, resulting in higher \textbf{label noise}. Second, federated parsing and a rich ontology admit the possibility of \textbf{multiple correct outputs}, where ``correct'' means acceptable downstream behavior. 

Testsets for supervised ML training generally presuppose a single correct output for a given item. Thus, component-level metrics may count many outputs as errors when they have no user-facing impact. Given a finite reannotation budget, the best system-level results are achieved by targeting the NLU errors that matter for the user (rather than spending resources to improve component-level metrics without improving system behavior). The challenge is to identify when divergent NLU outputs for the same input are correct; and incorporate that information into reannotation workflows.

As outlined in Figure \ref{fig-holistic-eval}, user interactions are replayed using different model versions (i.e. exploiting train-retrain variability). The baseline simply uses ground-truth for all components. The errors that lead to regression in user-facing response are subsequently prioritized for review and fix. In practice, only 10-50\% of reported model errors lead to user-facing errors. Therefore, this approach boosts engineering efficiency, and yields an end-to-end accuracy metric that is less dependent on the ontology (and therefore more comparable across time and divergent systems).

\section{Conclusions}

We have described an approach to the NLU system of a personal digitial assistant which can run entirely on a handheld device.  Key design decisions include a contextual reference resolution system and a limited notion of dialog state, hierarchical ontology-based meaning representation, sequence to sequence parsing producing tree-building instructions, and federated parsing to support personal vs. knowledge queries. We also developed a novel method for evaluating performance of the overall system by measuring user-facing failures due solely to NLU errors: this facilitates comparison between systems using different representations for user intents.

\bibliography{anthology,custom}

\begin{thebibliography}{25}
\expandafter\ifx\csname natexlab\endcsname\relax\def\natexlab#1{#1}\fi

\bibitem[{Balaraman et~al.(2021)Balaraman, Sheikhalishahi, and
  Magnini}]{balaraman-etal-2021-recent}
Vevake Balaraman, Seyedmostafa Sheikhalishahi, and Bernardo Magnini. 2021.
\newblock \href {https://aclanthology.org/2021.sigdial-1.25} {Recent neural
  methods on dialogue state tracking for task-oriented dialogue systems: A
  survey}.
\newblock In \emph{Proceedings of the 22nd Annual Meeting of the Special
  Interest Group on Discourse and Dialogue}, pages 239--251, Singapore and
  Online. Association for Computational Linguistics.

\bibitem[{Campagna et~al.(2017)Campagna, Ramesh, Xu, Fischer, and
  Lam}]{almond2017}
G~Campagna, R~Ramesh, S~Xu, M~Fischer, and M~S Lam. 2017.
\newblock Almond: The architecture of an open, crowdsourced,
  privacy-preserving, programmable virtual assistant.
\newblock In \emph{Proceedings of the 26th International Conference on World
  Wide Web}, pages 341--350.

\bibitem[{Carletta et~al.(1997)Carletta, Isard, Isard, Kowtko, Doherty-Sneddon,
  and Anderson}]{carletta-etal-1997-reliability}
Jean Carletta, Amy Isard, Stephen Isard, Jacqueline~C. Kowtko, Gwyneth
  Doherty-Sneddon, and Anne~H. Anderson. 1997.
\newblock \href {https://aclanthology.org/J97-1002} {The reliability of a
  dialogue structure coding scheme}.
\newblock \emph{Computational Linguistics}, 23(1):13--31.

\bibitem[{Cheng et~al.(2020)Cheng, Agrawal, Mart{\'\i}nez~Alonso, Bhargava,
  Driesen, Flego, Kaplan, Kartsaklis, Li, Piraviperumal, Williams, Yu,
  {\'O}~S{\'e}aghdha, and Johannsen}]{cheng-etal-2020-conversational}
Jianpeng Cheng, Devang Agrawal, H{\'e}ctor Mart{\'\i}nez~Alonso, Shruti
  Bhargava, Joris Driesen, Federico Flego, Dain Kaplan, Dimitri Kartsaklis, Lin
  Li, Dhivya Piraviperumal, Jason~D. Williams, Hong Yu, Diarmuid
  {\'O}~S{\'e}aghdha, and Anders Johannsen. 2020.
\newblock \href {https://doi.org/10.18653/v1/2020.emnlp-main.651}
  {Conversational semantic parsing for dialog state tracking}.
\newblock In \emph{Proceedings of the 2020 Conference on Empirical Methods in
  Natural Language Processing (EMNLP)}, pages 8107--8117, Online. Association
  for Computational Linguistics.

\bibitem[{Choe and Charniak(2016)}]{Choe2016ParsingAL}
Do~Kook Choe and Eugene Charniak. 2016.
\newblock Parsing as language modeling.
\newblock In \emph{Conference on Empirical Methods in Natural Language
  Processing}.

\bibitem[{Devlin et~al.(2019)Devlin, Chang, Lee, and
  Toutanova}]{devlin2019bert}
Jacob Devlin, Ming-Wei Chang, Kenton Lee, and Kristina Toutanova. 2019.
\newblock Bert: Pre-training of deep bidirectional transformers for language
  understanding.
\newblock In \emph{Proceedings of the 2019 Conference of the North American
  Chapter of the Association for Computational Linguistics: Human Language
  Technologies, Volume 1 (Long and Short Papers)}, pages 4171--4186.

\bibitem[{Gupta et~al.(2018)Gupta, Shah, Mohit, Kumar, and
  Lewis}]{gupta-etal-2018-semantic-parsing}
Sonal Gupta, Rushin Shah, Mrinal Mohit, Anuj Kumar, and Mike Lewis. 2018.
\newblock \href {https://doi.org/10.18653/v1/D18-1300} {Semantic parsing for
  task oriented dialog using hierarchical representations}.
\newblock In \emph{Proceedings of the 2018 Conference on Empirical Methods in
  Natural Language Processing}, pages 2787--2792, Brussels, Belgium.
  Association for Computational Linguistics.

\bibitem[{Ham et~al.(2020)Ham, Lee, Jang, and Kim}]{ham-etal-2020-end}
Donghoon Ham, Jeong-Gwan Lee, Youngsoo Jang, and Kee-Eung Kim. 2020.
\newblock \href {https://doi.org/10.18653/v1/2020.acl-main.54} {End-to-end
  neural pipeline for goal-oriented dialogue systems using {GPT}-2}.
\newblock In \emph{Proceedings of the 58th Annual Meeting of the Association
  for Computational Linguistics}, pages 583--592, Online. Association for
  Computational Linguistics.

\bibitem[{Henderson et~al.(2014)Henderson, Thomson, and
  Young}]{henderson-etal-2014-word}
Matthew Henderson, Blaise Thomson, and Steve Young. 2014.
\newblock \href {https://doi.org/10.3115/v1/W14-4340} {Word-based dialog state
  tracking with recurrent neural networks}.
\newblock In \emph{Proceedings of the 15th Annual Meeting of the Special
  Interest Group on Discourse and Dialogue ({SIGDIAL})}, pages 292--299,
  Philadelphia, PA, U.S.A. Association for Computational Linguistics.

\bibitem[{Jiao et~al.(2020)Jiao, Yin, Shang, Jiang, Chen, Li, Wang, and
  Liu}]{jiao-etal-2020-tinybert}
Xiaoqi Jiao, Yichun Yin, Lifeng Shang, Xin Jiang, Xiao Chen, Linlin Li, Fang
  Wang, and Qun Liu. 2020.
\newblock \href {https://doi.org/10.18653/v1/2020.findings-emnlp.372}
  {{T}iny{BERT}: Distilling {BERT} for natural language understanding}.
\newblock In \emph{Findings of the Association for Computational Linguistics:
  EMNLP 2020}, pages 4163--4174, Online. Association for Computational
  Linguistics.

\bibitem[{Kollar et~al.(2018)Kollar, Berry, Stuart, Owczarzak, Chung, Mathias,
  Kayser, Snow, and Matsoukas}]{kollar-etal-2018-alexa}
Thomas Kollar, Danielle Berry, Lauren Stuart, Karolina Owczarzak, Tagyoung
  Chung, Lambert Mathias, Michael Kayser, Bradford Snow, and Spyros Matsoukas.
  2018.
\newblock \href {https://doi.org/10.18653/v1/N18-3022} {The {A}lexa meaning
  representation language}.
\newblock In \emph{Proceedings of the 2018 Conference of the North {A}merican
  Chapter of the Association for Computational Linguistics: Human Language
  Technologies, Volume 3 (Industry Papers)}, pages 177--184, New Orleans -
  Louisiana. Association for Computational Linguistics.

\bibitem[{Li et~al.(2020)Li, Qu, and Haffari}]{Li2020ContextDS}
Zhuang Li, Lizhen Qu, and Gholamreza Haffari. 2020.
\newblock Context dependent semantic parsing: A survey.
\newblock \emph{ArXiv}, abs/2011.00797.

\bibitem[{Liu et~al.(2019)Liu, Gardner, Belinkov, Peters, and
  Smith}]{liu-etal-2019-linguistic}
Nelson~F. Liu, Matt Gardner, Yonatan Belinkov, Matthew~E. Peters, and Noah~A.
  Smith. 2019.
\newblock \href {https://doi.org/10.18653/v1/N19-1112} {Linguistic knowledge
  and transferability of contextual representations}.
\newblock In \emph{Proceedings of the 2019 Conference of the North {A}merican
  Chapter of the Association for Computational Linguistics: Human Language
  Technologies, Volume 1 (Long and Short Papers)}, pages 1073--1094,
  Minneapolis, Minnesota. Association for Computational Linguistics.

\bibitem[{Meng et~al.(2022)Meng, Dai, Wang, Wang, Wu, Jiang, and
  Liu}]{meng2022lexicon}
Xiaojun Meng, Wenlin Dai, Yasheng Wang, Baojun Wang, Zhiyong Wu, Xin Jiang, and
  Qun Liu. 2022.
\newblock Lexicon-injected semantic parsing for task-oriented dialog.
\newblock \emph{arXiv preprint arXiv:2211.14508}.

\bibitem[{Mrk{\v{s}}i{\'c} et~al.(2015)Mrk{\v{s}}i{\'c}, {\'O}~S{\'e}aghdha,
  Thomson, Ga{\v{s}}i{\'c}, Su, Vandyke, Wen, and
  Young}]{mrksic-etal-2015-multi}
Nikola Mrk{\v{s}}i{\'c}, Diarmuid {\'O}~S{\'e}aghdha, Blaise Thomson, Milica
  Ga{\v{s}}i{\'c}, Pei-Hao Su, David Vandyke, Tsung-Hsien Wen, and Steve Young.
  2015.
\newblock \href {https://doi.org/10.3115/v1/P15-2130} {Multi-domain dialog
  state tracking using recurrent neural networks}.
\newblock In \emph{Proceedings of the 53rd Annual Meeting of the Association
  for Computational Linguistics and the 7th International Joint Conference on
  Natural Language Processing (Volume 2: Short Papers)}, pages 794--799,
  Beijing, China. Association for Computational Linguistics.

\bibitem[{Mrk{\v{s}}i{\'c} et~al.(2017)Mrk{\v{s}}i{\'c}, {\'O}~S{\'e}aghdha,
  Wen, Thomson, and Young}]{mrksic-etal-2017-neural}
Nikola Mrk{\v{s}}i{\'c}, Diarmuid {\'O}~S{\'e}aghdha, Tsung-Hsien Wen, Blaise
  Thomson, and Steve Young. 2017.
\newblock \href {https://doi.org/10.18653/v1/P17-1163} {Neural belief tracker:
  Data-driven dialogue state tracking}.
\newblock In \emph{Proceedings of the 55th Annual Meeting of the Association
  for Computational Linguistics (Volume 1: Long Papers)}, pages 1777--1788,
  Vancouver, Canada. Association for Computational Linguistics.

\bibitem[{Rastogi et~al.(2019)Rastogi, Gupta, Chen, and
  Lambert}]{rastogi-etal-2019-scaling}
Pushpendre Rastogi, Arpit Gupta, Tongfei Chen, and Mathias Lambert. 2019.
\newblock \href {https://doi.org/10.18653/v1/N19-2013} {Scaling multi-domain
  dialogue state tracking via query reformulation}.
\newblock In \emph{Proceedings of the 2019 Conference of the North {A}merican
  Chapter of the Association for Computational Linguistics: Human Language
  Technologies, Volume 2 (Industry Papers)}, pages 97--105, Minneapolis,
  Minnesota. Association for Computational Linguistics.

\bibitem[{Ren et~al.(2019)Ren, Ni, and McAuley}]{ren-etal-2019-scalable}
Liliang Ren, Jianmo Ni, and Julian McAuley. 2019.
\newblock \href {https://doi.org/10.18653/v1/D19-1196} {Scalable and accurate
  dialogue state tracking via hierarchical sequence generation}.
\newblock In \emph{Proceedings of the 2019 Conference on Empirical Methods in
  Natural Language Processing and the 9th International Joint Conference on
  Natural Language Processing (EMNLP-IJCNLP)}, pages 1876--1885, Hong Kong,
  China. Association for Computational Linguistics.

\bibitem[{Rogers et~al.(2020)Rogers, Kovaleva, and
  Rumshisky}]{rogers-etal-2020-primer}
Anna Rogers, Olga Kovaleva, and Anna Rumshisky. 2020.
\newblock \href {https://doi.org/10.1162/tacl_a_00349} {A primer in
  {BERT}ology: What we know about how {BERT} works}.
\newblock \emph{Transactions of the Association for Computational Linguistics},
  8:842--866.

\bibitem[{Tseng et~al.(2021)Tseng, Bhargava, Lu, Moniz, Piraviperumal, Li, and
  Yu}]{tseng-etal-2021-cread}
Bo-Hsiang Tseng, Shruti Bhargava, Jiarui Lu, Joel Ruben~Antony Moniz, Dhivya
  Piraviperumal, Lin Li, and Hong Yu. 2021.
\newblock \href {https://doi.org/10.18653/v1/2021.naacl-main.265} {{CREAD}:
  Combined resolution of ellipses and anaphora in dialogues}.
\newblock In \emph{Proceedings of the 2021 Conference of the North American
  Chapter of the Association for Computational Linguistics: Human Language
  Technologies}, pages 3390--3406, Online. Association for Computational
  Linguistics.

\bibitem[{Wen et~al.(2017)Wen, Vandyke, Mrk{\v{s}}i{\'c}, Ga{\v{s}}i{\'c},
  Rojas-Barahona, Su, Ultes, and Young}]{wen-etal-2017-network}
Tsung-Hsien Wen, David Vandyke, Nikola Mrk{\v{s}}i{\'c}, Milica
  Ga{\v{s}}i{\'c}, Lina~M. Rojas-Barahona, Pei-Hao Su, Stefan Ultes, and Steve
  Young. 2017.
\newblock \href {https://aclanthology.org/E17-1042} {A network-based end-to-end
  trainable task-oriented dialogue system}.
\newblock In \emph{Proceedings of the 15th Conference of the {E}uropean Chapter
  of the Association for Computational Linguistics: Volume 1, Long Papers},
  pages 438--449, Valencia, Spain. Association for Computational Linguistics.

\bibitem[{Wu et~al.(2019)Wu, Madotto, Hosseini-Asl, Xiong, Socher, and
  Fung}]{wu-etal-2019-transferable}
Chien-Sheng Wu, Andrea Madotto, Ehsan Hosseini-Asl, Caiming Xiong, Richard
  Socher, and Pascale Fung. 2019.
\newblock \href {https://doi.org/10.18653/v1/P19-1078} {Transferable
  multi-domain state generator for task-oriented dialogue systems}.
\newblock In \emph{Proceedings of the 57th Annual Meeting of the Association
  for Computational Linguistics}, pages 808--819, Florence, Italy. Association
  for Computational Linguistics.

\bibitem[{Young(2014)}]{young-2014-keynote}
Steve Young. 2014.
\newblock \href {https://doi.org/10.3115/v1/W14-4301} {{K}eynote: Statistical
  approaches to open-domain spoken dialogue systems}.
\newblock In \emph{Proceedings of the 15th Annual Meeting of the Special
  Interest Group on Discourse and Dialogue ({SIGDIAL})}, page~1, Philadelphia,
  PA, U.S.A. Association for Computational Linguistics.

\bibitem[{Young et~al.(2013)Young, Gasic, Thomson, and
  Williams}]{Young2013POMDPBasedSS}
Steve~J. Young, Milica Gasic, Blaise Thomson, and J.~Williams. 2013.
\newblock Pomdp-based statistical spoken dialog systems: A review.
\newblock \emph{Proceedings of the IEEE}, 101:1160--1179.

\bibitem[{Yu et~al.(2018)Yu, Lin, Shen, Yang, Lu, Bansal, and Berg}]{MAttNet}
Licheng Yu, Zhe Lin, Xiaohui Shen, Jimei Yang, Xin Lu, Mohit Bansal, and
  Tamara~L. Berg. 2018.
\newblock \href {http://arxiv.org/abs/arXiv:1801.08186} {Mattnet: Modular
  attention network for referring expression comprehension}.
\newblock ArXiv:1801.08186.

\end{thebibliography}
\bibliographystyle{acl_natbib}

\end{document}